\documentclass{llncs}
\usepackage{subcaption}
\usepackage{array}
\usepackage{amsfonts, amssymb}
\usepackage{graphicx,amsmath}
\usepackage{hyperref}
\usepackage{caption}
\usepackage{comment}
\usepackage{soul}
\usepackage{makeidx}  
\usepackage{lineno}
\DeclareMathOperator*{\argmin}{arg\,min}
\begin{document}

\mainmatter              
\title{Extraction of airway trees using multiple hypothesis tracking and template matching}
\titlerunning{MHT}  
%
\author{Raghavendra Selvan\inst{1}, Jens Petersen\inst{1}, Jesper H. Pedersen\inst{2} \and
Marleen de Bruijne\inst{1,3}}
\authorrunning{Raghavendra Selvan et al.} 
%
\tocauthor{Raghavendra Selvan, Jens Petersen, Jesper Pedersen, Marleen de Bruijne}
\institute{Department of Computer Science, University of Copenhagen, Denmark
\and
Department of Cardio-Thoracic Surgery RT, Rigshospitalet, University Hospital of Copenhagen, Denmark
\and
Departments of Medical Informatics and Radiology, Erasmus MC Rotterdam, The Netherlands
\\
\email{raghav@di.ku.dk}\\ 
}

\maketitle              

\begin{abstract}
Knowledge of airway tree morphology has important clinical applications in diagnosis of chronic obstructive pulmonary disease. We present an automatic tree extraction method based on multiple hypothesis tracking and template matching for this purpose and evaluate its performance on chest CT images. The method is adapted from a semi-automatic method devised for vessel segmentation. Idealized tubular templates are constructed that match airway probability obtained from a trained classifier and ranked based on their relative significance. Several such regularly spaced templates form the local hypotheses used in constructing a multiple hypothesis tree, which is then traversed to reach decisions. The proposed modifications remove the need for local thresholding of hypotheses as decisions are made entirely based on statistical comparisons involving the hypothesis tree. The results show improvements in performance when compared to the original method and region growing on intensity images. We also compare the method with region growing on the probability images, where the presented method does not show substantial improvement, but we expect it to be less sensitive to local anomalies in the data.
\keywords{airways, tracking, segmentation, computed tomography}
\end{abstract}

\section{Introduction}

Segmentation of airway trees in Computed Tomography (CT) images has multiple important clinical applications. One such prominent application is in the diagnosis of chronic obstructive pulmonary disease, which is characterised by destruction of lung tissue and changes to the morphology of the airway tree. 

Several existing methods attempt to solve the task of airway segmentation, invariably, by balancing a trade-off between sensitivity and specificity, consistency on different data sets, the extent of user interaction and running time. In a comparitive study of airway segmentation methods, published in the EXACT'09 study~\cite{exact}, 15 different methods were evaluated encompassing most of the state-of-the-art. One important take-away from this study is that there is scope for improvement in the aforementioned trade-offs. Thus leaving room for an automatic airway segmentation method that is accurate,  robust and fast.

\begin{figure}[t]
\begin{subfigure}{0.5\textwidth}
\centering
\begin{minipage}{0.5\textwidth}
\includegraphics[width=1.2\textwidth]{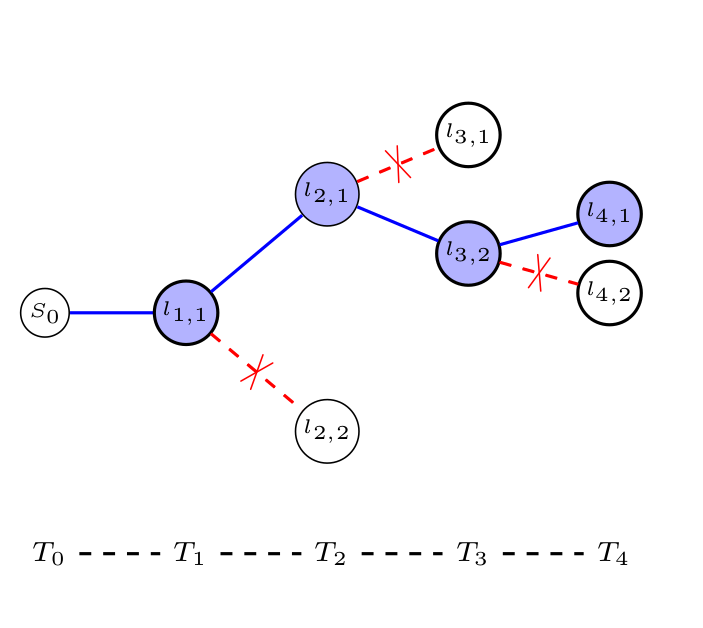}
\end{minipage}
\caption{Instantaneous decisions}
\label{fig:singleHyp}
\end{subfigure}
\hspace{0.25cm}
\begin{subfigure}{0.5\textwidth}
\centering
\begin{minipage}{0.5\textwidth}
\includegraphics[width=1.2\textwidth]{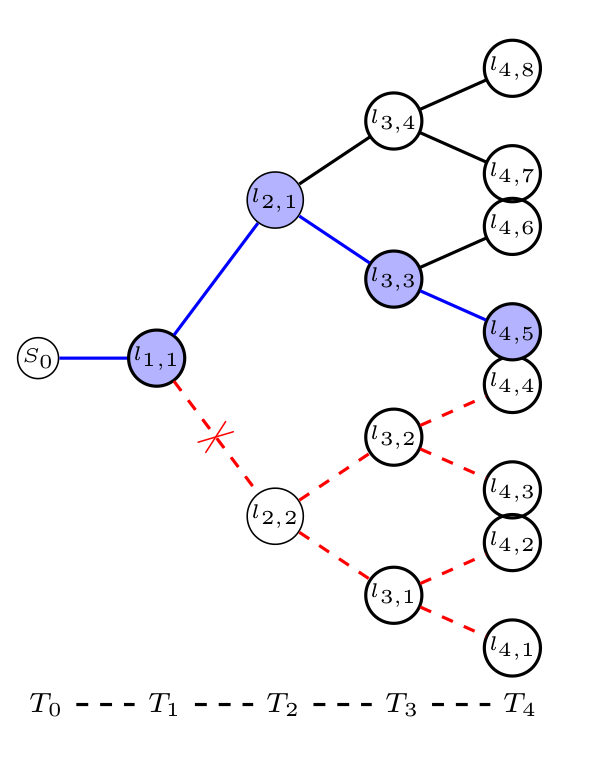}
\end{minipage}
\caption{Deferred decisions}
\label{fig:mht}
\end{subfigure}
\caption{Tree view of single hypothesis and multiple hypothesis tracking methods. Hypothesis trees are shown for tracking steps $T_0,\dots,T_4$ with an initial state $S_0$ and local hypothesis $i$ at step $t$ is denoted as $l_{t,i}$. In case of instantaneous decisions in Figure~\ref{fig:singleHyp}, only the best local hypothesis at each instant is retained (marked in blue). For the case of MHT in Figure~\ref{fig:mht}, the decision at $T_2$ is based on the best global hypothesis at step $T_4$, marked with blue edges. Once the decision at $T_2$ has been made, hypotheses that are not children of the best node are discarded, shown in red.} 
\label{fig:comparison}
\end{figure}

In this work, we explore the possibility of using multiple hypothesis tracking (MHT)~\cite{reid,blackman} -- a well known decision making paradigm in multi-target tracking -- for segmenting airway trees. 
Compared to conventional tracking methods where decisions are made instantaneously, in MHT the decisions are deferred to a future tracking step controlled by the search depth. 
From any given step, all possible solutions are maintained up to search depth in the form of a hypothesis tree, and the decision at each step is made by tracing back the best hypothesis at the end of the hypothesis tree. The hypotheses generated at every tracking step in an MHT tree are termed as local hypotheses, and sequence of local hypotheses across tracking steps from the last common root node are the global hypotheses. The quality of global hypotheses is decided based on the average score of their local hypotheses, referred to as the global score. The feature of deferring decisions in MHT can considerably improve segmentation results, as it incorporates additional global information, when compared to methods that rely on making instantaneous decisions such as region growing-based methods which are often used in airway segmentation~\cite{mori,sonka,vessel}. The concept of instantaneous decisions are illustrated along with deferred decisions using hypothesis trees in Figure~\ref{fig:comparison}. 
%
%




In a previous work ~\cite{friman}, MHT was used along with template matching for segmenting small and bright tubular structures from a dark background in a semi-automatic manner. This method starts tracking from manually placed seed points and constructs an MHT tree comprising of local hypotheses, which are assigned scores based on the contrast signal-to-noise ratio (SNR) of the tubular template in the image data. A local hypothesis threshold is used to discard low quality local hypotheses. Using these thresholded local hypotheses, the MHT tree is constructed by retaining global hypotheses that are above a certain global threshold. When tracking of a branch stops, further seed points are placed by the user and tracking is restarted. The method in~\cite{friman} has certain limitations when applied to segmenting trees; most importantly that the local and global threshold parameters used are scale sensitive. There are also issues in the way the method handles bifurcations, which is using a basic approach based on clustering and user intervention. 

The presented method, which is based on~\cite{friman}, overcomes the above mentioned limitations by introducing statistical ranking of local hypotheses, which is described in Section~\ref{sec:ranking}. A favourable feature of the modified method is that the local threshold parameter is no longer required and the global threshold parameter does not depend on scale of the structures. 
In the remainder of the paper, we describe the proposed method and the differences with the original MHT method in~\cite{friman}.  We evaluate the modified method on computed tomography (CT) data from the Danish Lung Cancer Screening Trials~(DLCST)~\cite{dlcst}.  We compare the modified method with the original method in~\cite{friman}, and also with region growing on intensity and probability images obtained from a trained classifier.

\section{Method}
The method presented in this paper is based on the work in~\cite{friman}. In this section, we present aspects of the original method that are retained in our modified method, describe the limitations in it and present the modifications that enable our method to be used for extraction of airway trees. For convenience, the method in~\cite{friman} will be addressed as the original MHT method and our's as the modified MHT method. 

\subsection{The original MHT method}
\label{subsec:template}

In the original MHT method in\cite{friman}, long tubular structures are modelled as a cascading sequence of tubular segments and are tracked using MHT and template matching. The image data is tested in sequential steps (of the image volume and not time) for the presence of tubular structures by performing matched filtering with a template. The distance between successive tracking steps is based on the previous radius of the template and a scalar factor, referred to as the step length factor. From every tracking step, a number of tubular templates are placed at different angles with respect to the current direction of the branch to cover a certain volume of the data; this volume is a sphere segment and is controlled by the maximum search angle parameter. Each of these template predictions form the local hypothesis used to construct the MHT tree and are assigned a score based on the response of template matching.

To perform template matching, first an idealized 3-D tubular template is modelled which maps variations of the intensity to values between $0$ and $1$, i.e, $T(\mathbf{x};\mathbf{x}_0, \mathbf{\hat{v}}, r ):\mathbb{R}\rightarrow~[0,1]$, given as,
\begin{equation}
T(\mathbf{x};\mathbf{x}_0, \mathbf{\hat{v}}, r ) = \frac{r^{\gamma}}{(d^2(\mathbf{x};\mathbf{x}_0,\mathbf{\hat{v}}))^{\gamma/2}+r^{\gamma}},
\end{equation}
where $r$ is the radius of the template, $\gamma$ controls the steepness of the profile function and $d^2(\mathbf{x};\mathbf{x}_0,\mathbf{\hat{v}})$ is the squared distance between a point $\mathbf{x} \in \mathbb{R}^3$ and axis of the tubular structure along a unit vector $\mathbf{\hat{v}}$ through the point $\mathbf{x}_0$, which is the centre of the tubular template.

Next, the image neigbourhood $I(\mathbf{x})$ is related to the template function, contrast, $k$, and the background intensity, $m$, as
\begin{equation}
I(\mathbf{x}) = kT(\mathbf{x};\mathbf{x}_0, \mathbf{\hat{v}}, r ) + m + \epsilon(\mathbf{x})
\label{eq:image}
\end{equation}
where $\epsilon(\mathbf{x})$ captures the noise due to background structures. The relation in~\eqref{eq:image} models bright tubular structures of contrast $k$ in a dark background of mean intensity $m$ with uncertainties due to interfering structures. The tubular template fitting is carried out by solving the following weighted least-squares problem:
\begin{equation}
\argmin_{k,m,r,\mathbf{x}_0,\mathbf{\hat{v}}} || \mathbf{W}(r,\mathbf{x}_0,\mathbf{\hat{v}})[kT(\mathbf{x}_0, \mathbf{\hat{v}}, r ) + m\mathbf{1}_n - \mathbf{I}] || ^2.
\label{eq:min}
\end{equation}
The weighting matrix $\mathbf{W}(r,\mathbf{x}_0,\mathbf{\hat{v}})$ has diagonal entries corresponding to an asymmetric Gaussian centered at $\mathbf{x}_0$ and is used to localize the fitting procedure. We refer to the original paper~\cite{friman} for details on the weighting matrix and the procedure to solve the minimization problem, which is carried out using the Levenberg-Marquardt algorithm. 

\begin{figure}[t]
\includegraphics[width=0.95\textwidth]{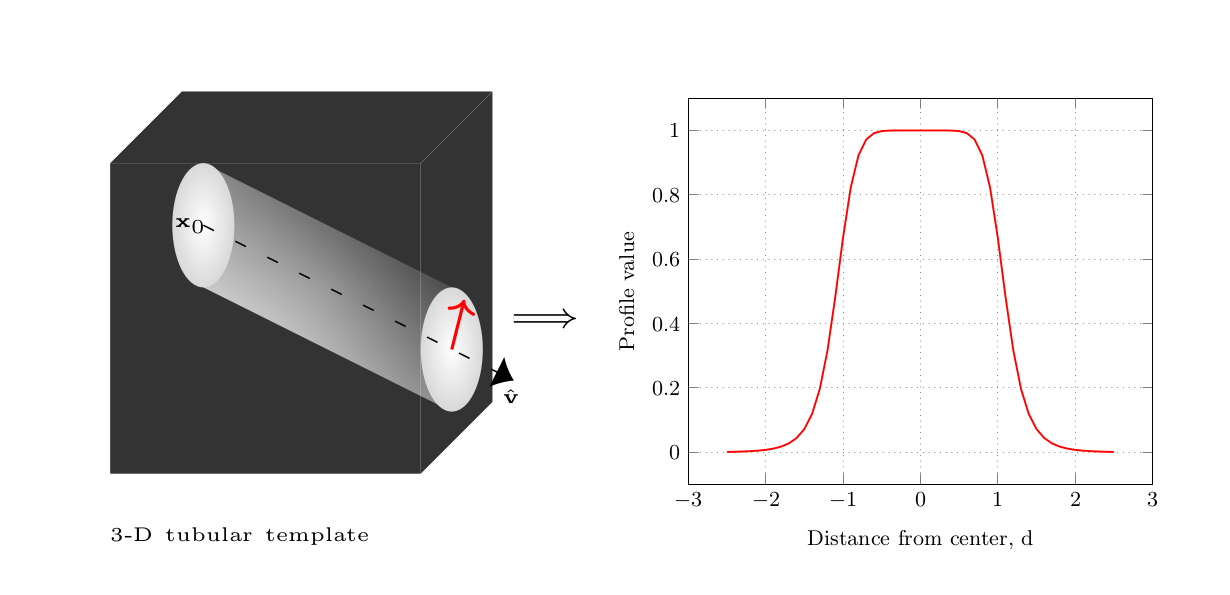}
\caption{ 3D tubular template of radius $r$, with center at $\mathbf{x}_0$ along the direction $\mathbf{\hat{v}}$. Intensity profile at a crossection is shown on the right.}
\label{fig:template}
\end{figure}

By solving the minimization problem in~\eqref{eq:min}, a tubular template is obtained -- with estimates of $(r,\mathbf{x}_0,\mathbf{\hat{v}})$ -- which fits the data in the neighbourhood $I(\mathbf{x})$ along with estimates of the image parameters -- $k, m$. The template fitting for a local hypothesis $i$ is quantified by computing the Student distribution's t-statistic, $l_i$,
\begin{equation}
l_i \triangleq \frac{k-m}{\text{std(k)}},
\label{eq:local}
\end{equation}
which gives a measure of how significantly different the bright tubular structure is from the background, or the contrast SNR. Local hypotheses below a threshold, referred to as the pruning threshold in~\cite{friman}, are discarded. 

In the original MHT method, a hypothesis tree of a predetermined search depth, $d_s$ is constructed using the local hypotheses. 
The decision of which of the global hypotheses to retain and which to discard is made based on the global hypothesis score, which for any global hypothesis of depth $d_l$ is
\begin{equation}
s_g = \frac{\sum_{i=1}^{d_l} l_{i}}{d_l}.
\label{eq:global}
\end{equation}
This score is compared with a global threshold and all hypotheses that do not exceed this threshold are discarded. Note that $d_l \leq d_s$, meaning hypotheses that are smaller than the depth of the MHT tree also have a global~score.

\subsection{Modifications to the original MHT method}
\label{sec:ranking}

The original MHT method was devised as an interactive method to track small bright structures in a dark background. It is not immediately applicable for the automatic extraction of airway trees from CT images. In this section, we elaborate on some of the limitations of the original MHT method in this respect and propose modifications to overcome them.

\subsubsection{Application to airway trees}

The intensity profile function of the original MHT method was designed to detect bright tubular structures in a dark background, however, airways in CT images appear as dark tubular shapes with bright walls in a dark background. Voxel classifiers trained to detect airway voxels have previously been shown to work better than intensity alone~\cite{vessel}, and the probabilistic outputs are a good fit with the intensity profile function of the original MHT method. We therefore chose to use a trained classifier and obtain probability images as described in Section~\ref{sec:probImages}.



\subsubsection{Varying dimensions}

An important factor to consider when tracking airways is the range of dimensions in which they appear. The radius of visible airways roughly range from 1 to 10 mm and this scale difference has a significant impact on the local hypothesis scores in~\eqref{eq:local}. As the local and global threshold parameters in the original MHT method are scale dependent, tuning these parameters to function well across airways of all dimensions poses a problem. 
Parameters that are tuned for structures of larger radii discard branches of lower radii. On the other hand, when the parameters are tuned to allow smaller branches, spurious branches of larger radii appear in the segmentation. In both cases scale-dependence of the local and global hypothesis thresholds is a limiting factor when segmenting structures of varying dimensions. 

The strategy we devise to overcome the scale dependence of local hypothesis scores is to rank the local hypotheses based on the score in~\eqref{eq:local}, mapping them to $[0,1]$. 
This scheme of relative scoring of hypotheses, where the best local hypothesis is assigned $1$, at each tracking step does not use the scale information directly and immediately alleviates the problem of scale dependence. Also, ranking of local hypotheses  captures their local significance or likelihood at each step. 

As a further simplification of the original method, we remove the threshold on local hypothesis scores and propagate all local hypotheses with their relative scores. Only a global hypothesis threshold is used which has a probabilistic interpretation to it. For instance, if the search depth is 10, a global hypothesis threshold of 0.6 would require the global hypotheses to have been the best locally at least in 6 of the 10 tracking steps, or similar. The probabilistic global hypothesis score is independent of the scale of the structures. Further, the use of ranking also makes the global threshold less dependent on the specific measure used to fit local hypotheses like the contrast SNR and can be replaced with any other fitness measure that can capture the relative quality of the local hypotheses.
One concern that arises when the scores in~\eqref{eq:local} are discarded for relative ranking is the chance of bad hypotheses performing well when ranked amongst equally bad counterparts.  A strength of the deferred decision in an MHT tree is that such hypothesis are unlikely to be accepted, as they would have to perform well across multiple steps. 

\subsubsection{Handling branching} Another factor to consider when tracking airway trees is the handling of branching. In~\cite{friman}, there is only a brief discussion about how bifurcations are handled. To detect branching, hypotheses are clustered into two separate clusters based on their spatial location at every step. The most significant leaf within each cluster is then determined to be a new branch.

Once a bifurcation is detected, there are several ways to proceed. As the tracking happens per branch, new seed points can be added from the detected points of bifurcation and tracking can be restarted. This entails rebuilding of the MHT tree from each of the new seed points. The implementation of the original MHT method, available as a MeVisLab~\footnote{\url{http://mevislab.de/}} module, follows this strategy. This strategy has a negative consequence of discarding the information aggregated until the step before branching. That is, the history of the evolution of the branch up to branching is abandoned when tracking is restarted from two new seed points. 

A further consequence of restarting tracking from branching points is that the new MHT trees can have shallow hypotheses. 
This causes low quality hypotheses to be accepted because as seen from equation~\eqref{eq:global}, the global hypothesis score per branch is computed by averaging over the leaf depth and not the search depth. 

Another strategy that does not rely on rebuilding the MHT tree is to simply save the MHT tree at branching and resume tracking from each of the newly detected branching points separately. By resuming tracking from both branching points with the history of the parent branch we do not throw away any information. In the modified MHT method, this strategy is used. Both seed points inherit the MHT tree at branching  and this simple transfer of history does not require rebuilding of the MHT tree but makes full use of it to make better decisions across bifurcations.

\section{Experiments and results}
\label{sec:results}

All experiments were conducted on a subset of 32 CT scans from the DLCST~\cite{dlcst} dataset. The images have a slice spacing of 1 mm and in-plane resolution between 0.72 to 0.78 mm. The images were randomly divided into training and test sets comprising of 16 images each. Reference segmentations obtained using manual annotations are available for these images and have been used as the ground truth to compare the performance. 
The modified MHT method was compared with the original MHT method in~\cite{friman}, region growing on intensity and region growing on probability.

\subsection{Pre-processing of data}
\label{sec:probImages}

\begin{figure}[h]
\centering
\begin{tabular}{c c c}
\includegraphics[height=3.25cm]{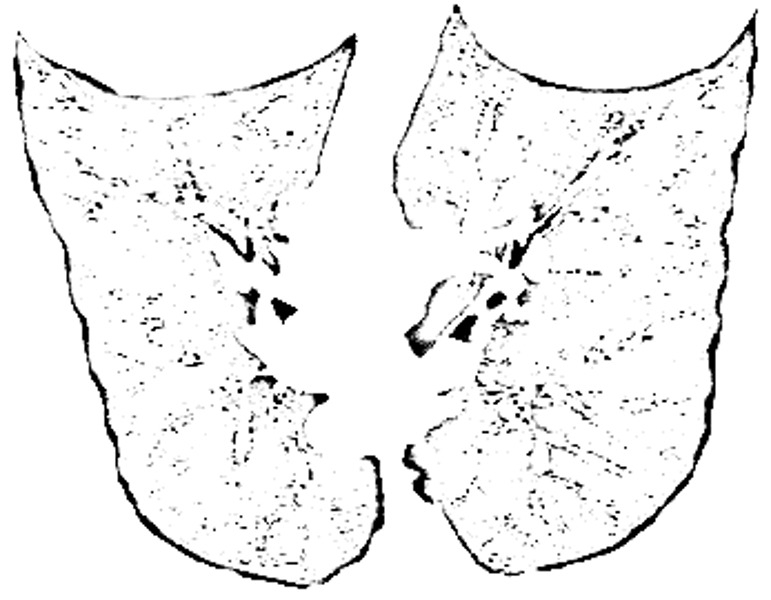} & \includegraphics[height=3.25cm]{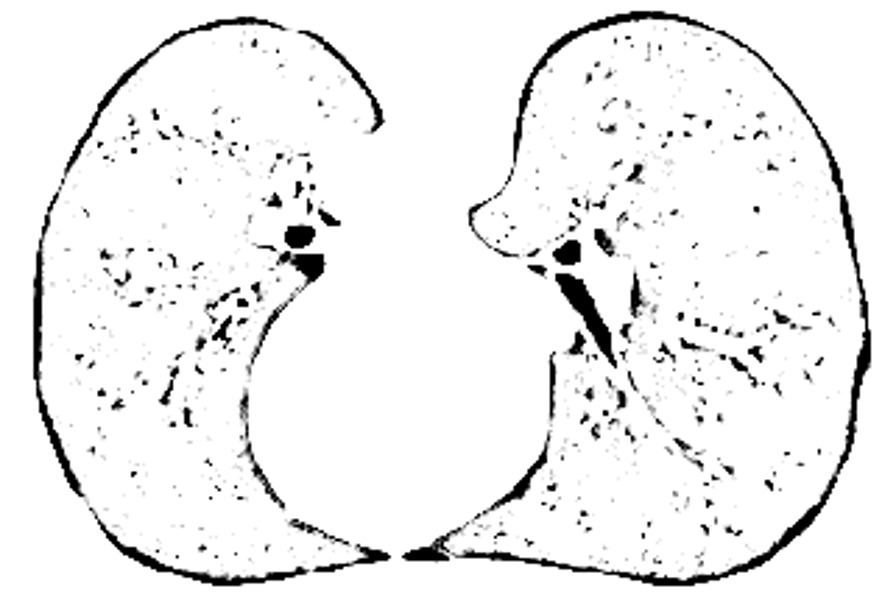} & \includegraphics[height=3.25cm]{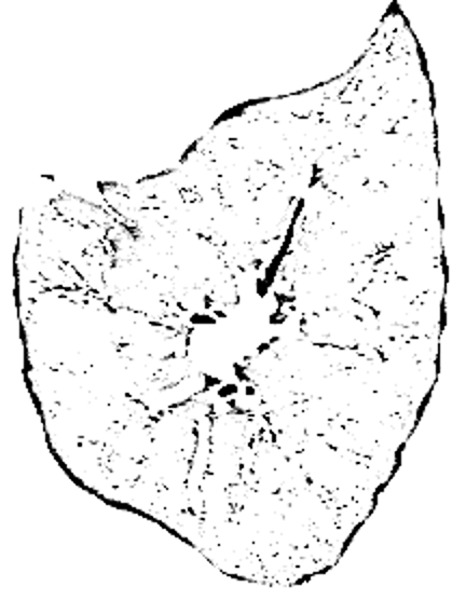}
\end{tabular}

\caption{Coronal, axial and sagital views of the probability image obtained from the trained classifier. Darker regions correspond to high probability, and hence likely airway regions.}
\label{fig:prob}
\end{figure}

The  CT images were pre-processed and converted into probability images using a k nearest neighbour (KNN)~\cite{knn} to match the profile function described in Section~\ref{subsec:template}. The KNN classification was performed using a neighbourhood of $K=21$ and multi-scale features using the method described in~\cite{vessel}. Thus obtained probability images have probability close to 1 in regions that are classified to be inside the airways and close to 0 outside. An example probability image is shown in Figure~\ref{fig:prob}. The noise observed is due to the acquisition, interfering vessels, lung tissue etcetera.
\subsection{Experiment set-up and parameter tuning}

The objective of the experiments is to evaluate the quality of the segmentation results obtained from the modified MHT method with respect to the ground truth and to compare it with the original MHT method, region growing on intensity and probability images. 

Parameters of all the methods were tuned on the training set to minimize the error measure defined in Section~\ref{subsec:error}. Using these tuned parameters, all methods were evaluated on an independent test set. The original MHT method has numerous tunable parameters. The first set of parameters are related to the tubular template: minimum and maximum allowable radii of the templates, scaling factors for weighting function and step length. Second category of parameters are related to the MHT: local and global hypothesis thresholds,  number of local hypotheses, maximum search angle and search depth. Tuning both categories of parameters for the training set is cumbersome, so some of the parameters were fixed based on the morphology of the airway trees. For instance, the minimum and maximum radii were fixed to 1 mm and 10 mm respectively, and the search depth was fixed to 6  based on initial training set experiments. Only the remaining parameters related to the MHT were tuned. Due to the scale independence introduced by the statistical ranking of local hypotheses, the modified MHT method on the other hand does have the local hypothesis threshold as a parameter. The set of parameters obtained by tuning on the training set for both MHT methods are presented in Table~\ref{tab:param}. Both the region growing methods only have the threshold parameter to be tuned. For the case of region growing on intensity of CT images, the threshold was found to be -995 HU, which is a really low threshold caused by the leakage of three images in the training set at any higher thresholds. The threshold for the case of probability images was found to be 0.5.
\newcolumntype{K}[1]{>{\centering\arraybackslash}p{#1}}
\vspace{-1cm}
\begin{table*}[ht!]
\caption{ Parameters for the MHT methods after tuning on the training set}
\label{tab:param}
\begin{center}
\small
    \begin{tabular}{| l | K{2cm} | K{1.75cm}|  K{1.75cm}| K{1cm}| K{1.5cm}| K{1.75cm}|}
    \hline
      & Weight window factor & Step Length Factor & Max. Search ang.(deg) & No. of angles & Loc. hyp. thresh. & Global hyp. thresh. \\ \hline
    Org. MHT & 3 & 1.5 & 60 & 3 & 2 & 4 \\ \hline
    Mod. MHT & 1 & 1.1 & 70 & 2 & -- & 0.7 \\ \hline
    \end{tabular}
\end{center}
\end{table*}

\subsection{Error measure}
\label{subsec:error}
We use a distance measure based on comparing the centerlines obtained from each of the methods with the centerlines of the reference segmentation to quantify the performance~\cite{cl}. The error measure is a sum of two distances, 
\begin{equation}
d_{err} = w\frac{\sum_{i=1}^{n_{op}} \min d_E(c_i-C_{ref})}{n_{op}}
+ (1-w)\frac{\sum_{j=1}^{n_{ref}} \min d_E(c_j-C_{op})}{n_{ref}}
\label{eq:error}
\end{equation}
where $C_{ref}$, $C_{op}$ are centerlines of the reference and output segmentation results, respectively comprising of $n_{ref}$ and $n_{op}$ number of points, $c_i, c_j \in \mathbb R^n$ are the individual points in the centerlines. The Euclidean distance is denoted as $d_E$ and $w$ is the weighting on the first distance, such that $ 0 \leq w \leq 1$. Notice that the first term in~\eqref{eq:error} captures the distance between the two centerlines due to false positives, whereas the second term captures the distance due to false negatives. We improve both the false positive and false negative performance, equally, by minimizing the error with $w=0.5$. Depending on the application the weight can be modified to obtain a desirable measure that reflects the sensitivity or specificity needs.

\begin{figure}[t]
\centering
\includegraphics[width=0.8\textwidth]{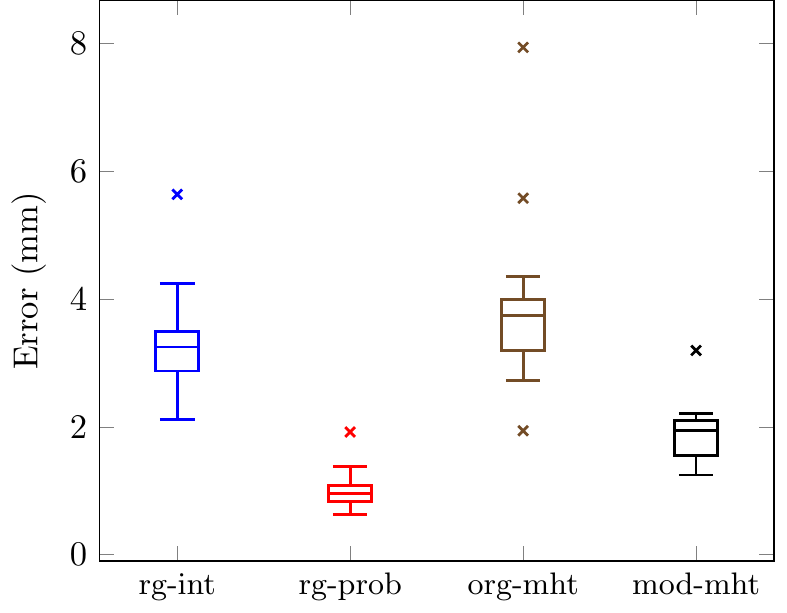}
\caption{ Performance comparison of the modified MHT (mod-MHT) method with the original MHT (org-MHT), region growing on intensity (rg-int) and region growing on probability (rg-prob) using standard box plots.}
\label{fig:errPlot}
\end{figure}

\subsection{Results}

\begin{figure}
\begin{center}
\begin{tabular}{c c c c}
      {\includegraphics[height=2.25cm]{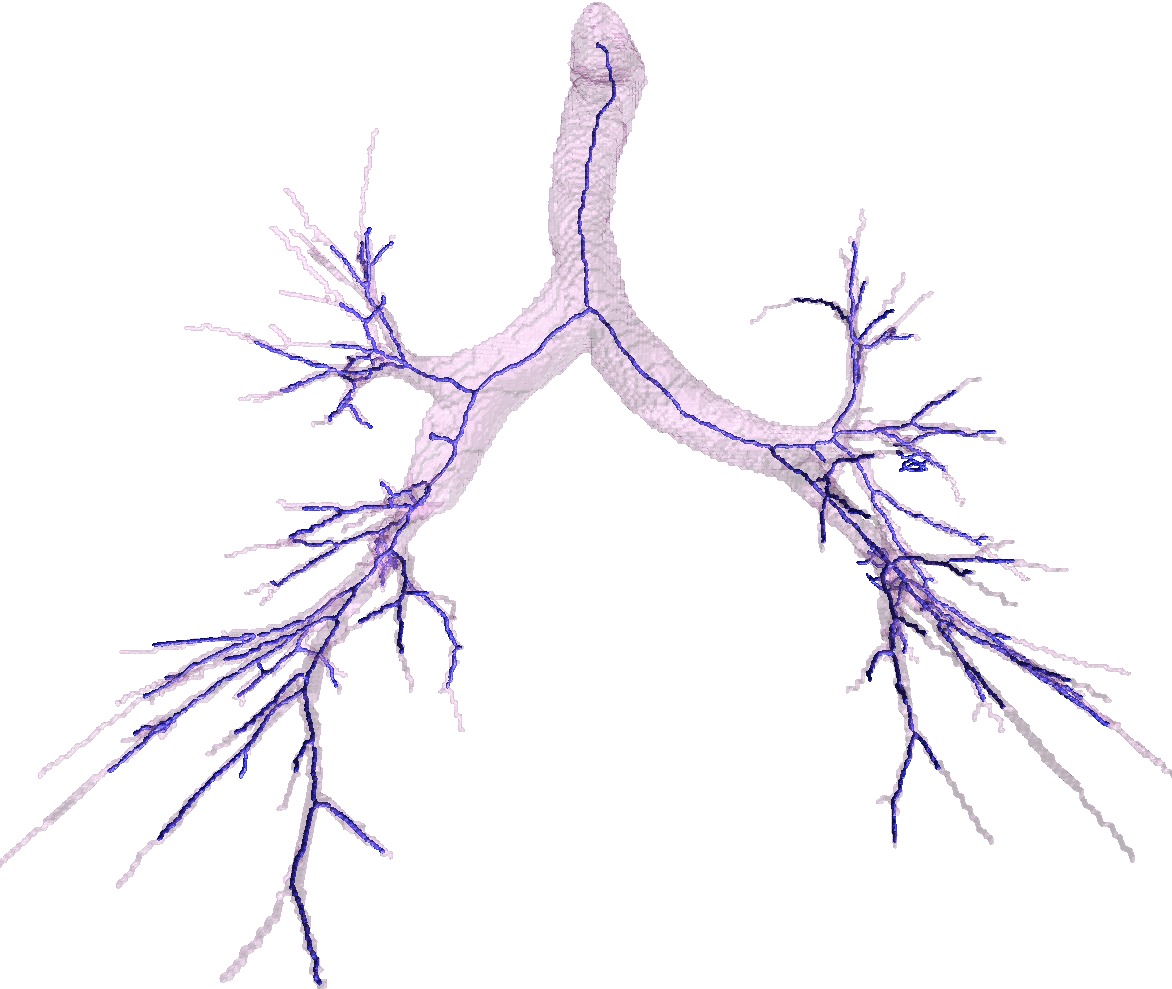}} &
      {\includegraphics[height=2.25cm]{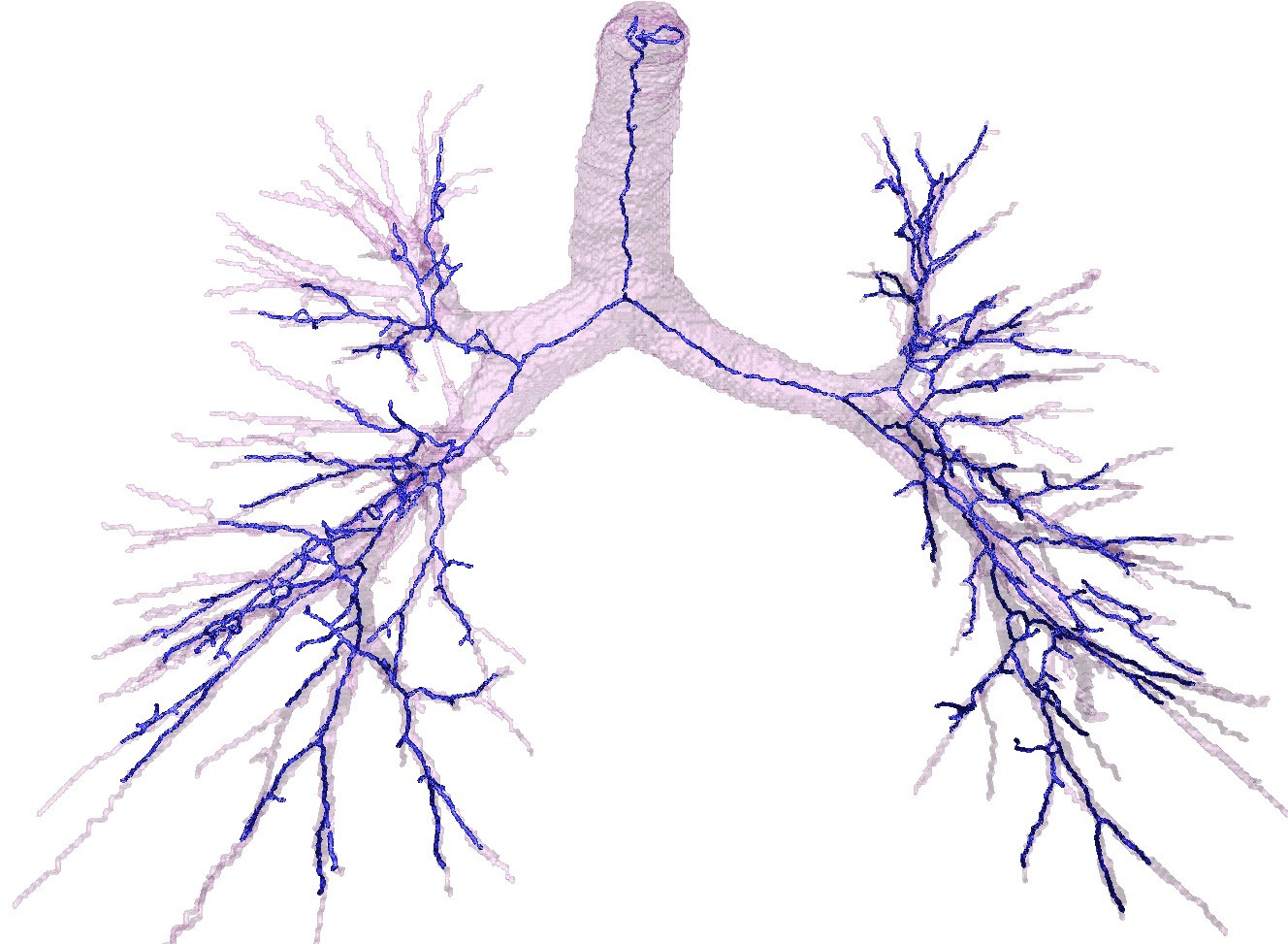}} & {\includegraphics[height=2.25cm]{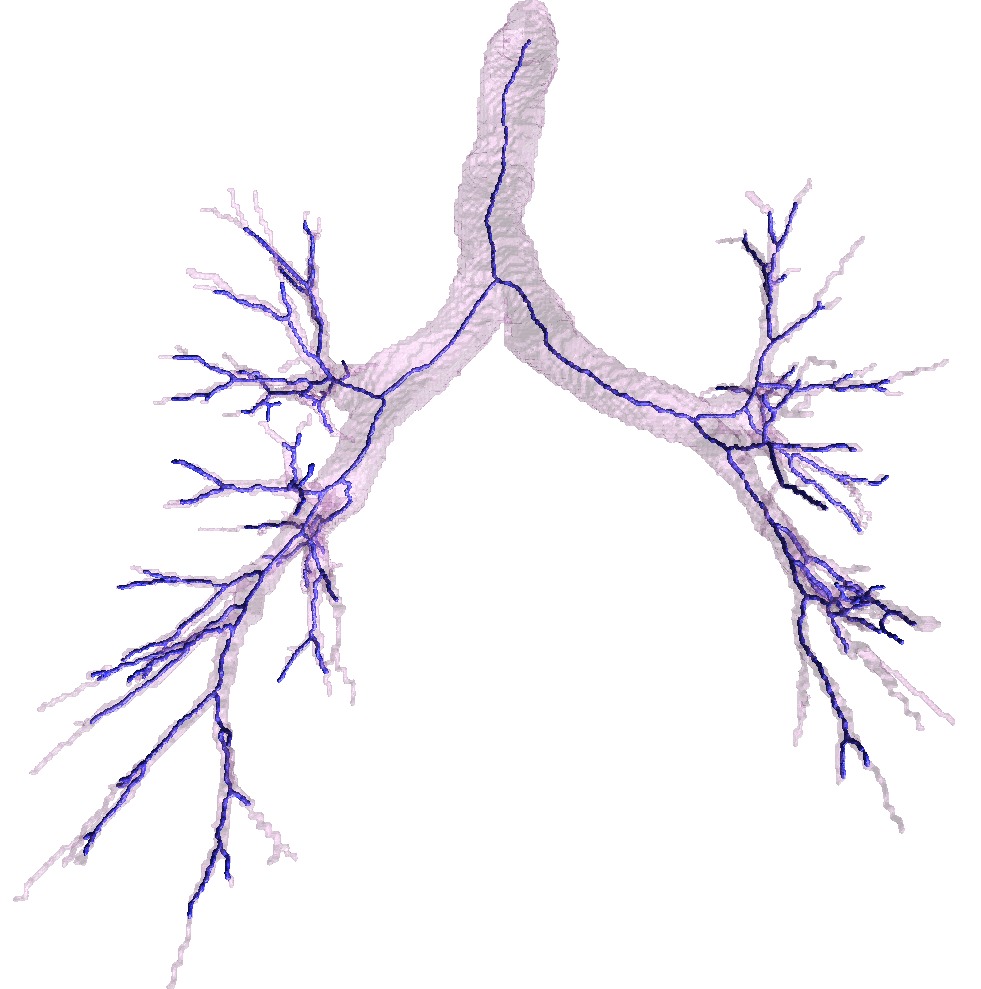}} & {\includegraphics[height=2.25cm]{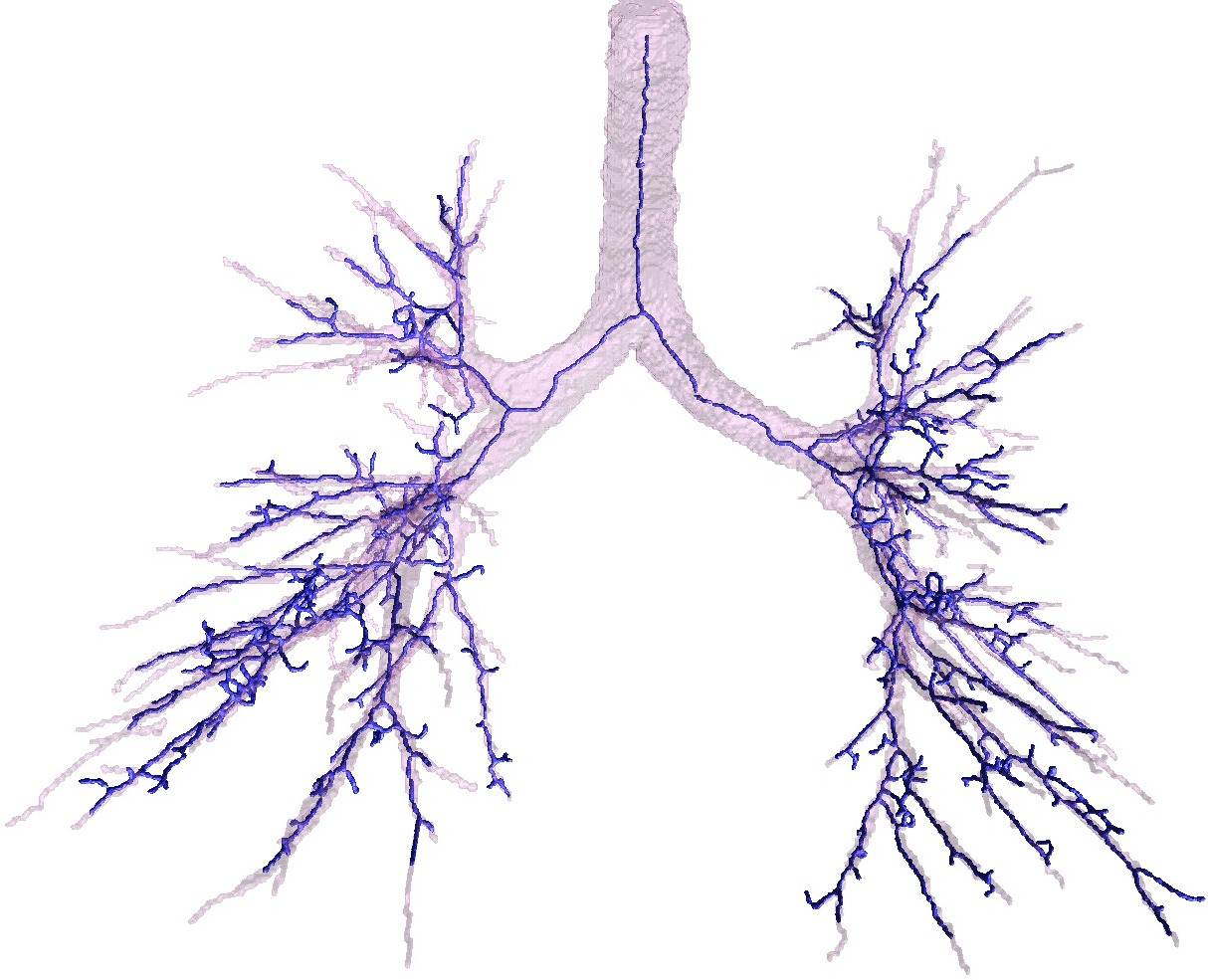}}\\ 
\vspace{0.3cm}
      (1) & (2) & (3) & (4) \\
      {\includegraphics[height=2.25cm]{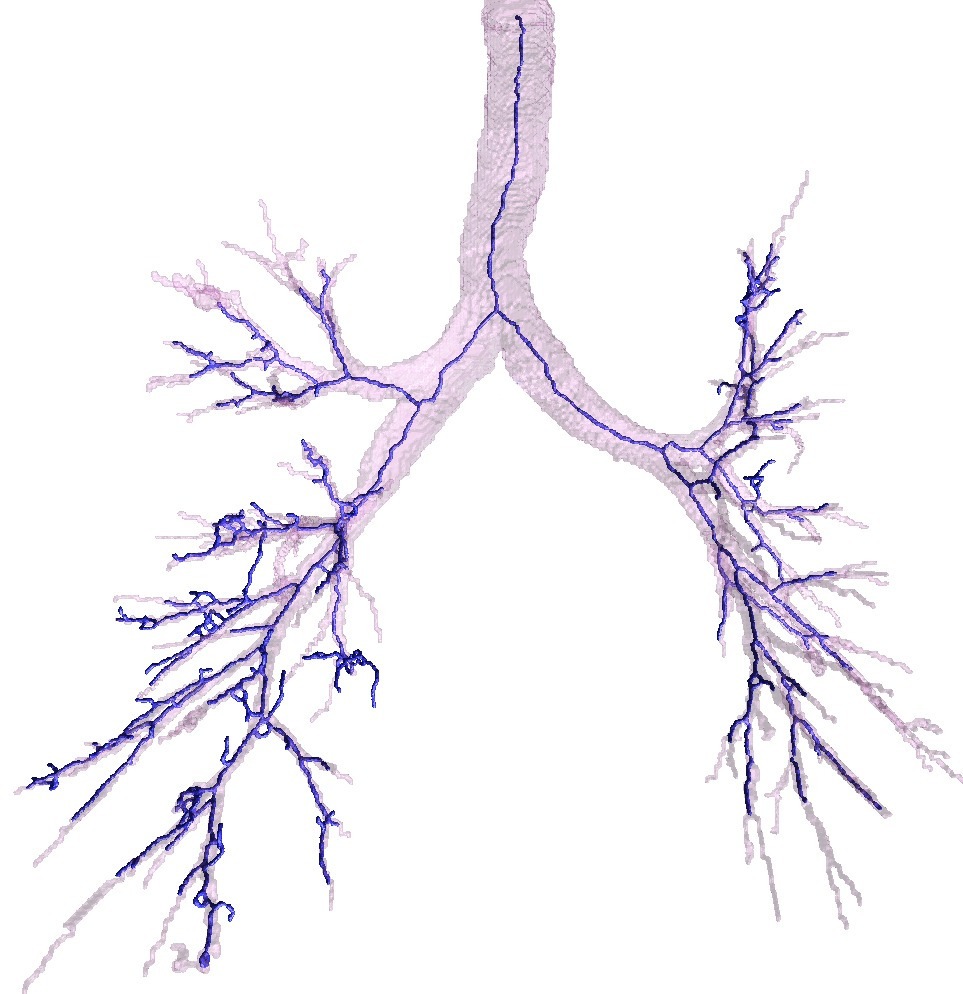}} &
      {\includegraphics[height=2.25cm]{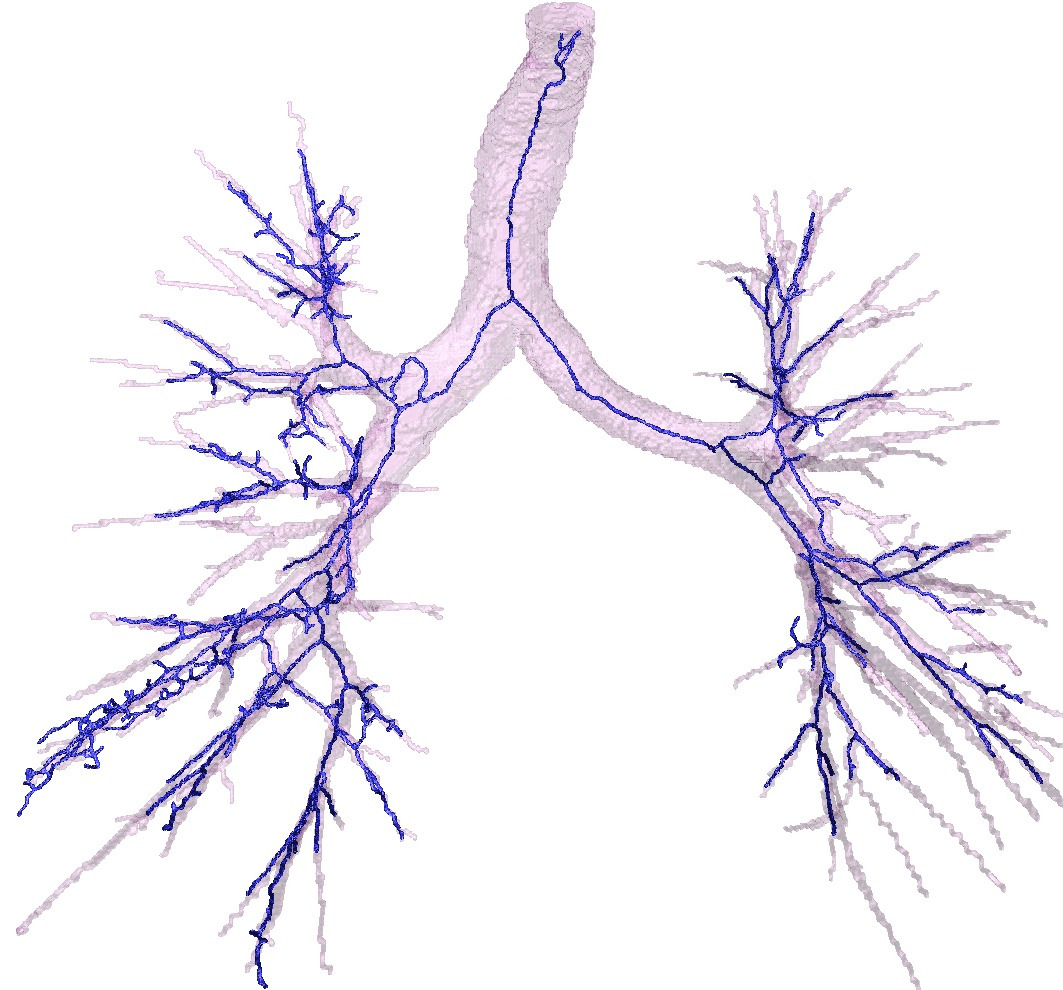}} & {\includegraphics[height=2.25cm]{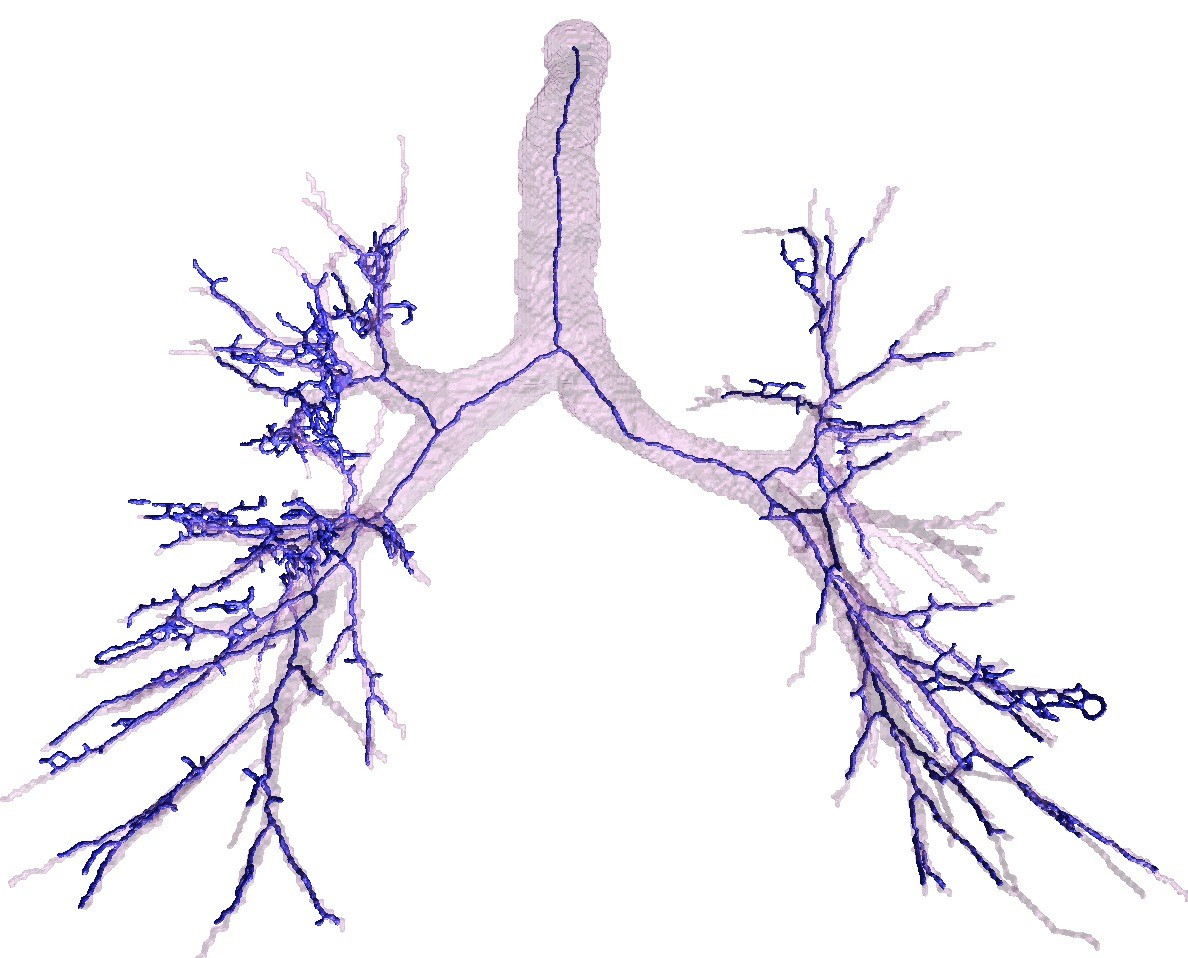}} & {\includegraphics[height=2.25cm]{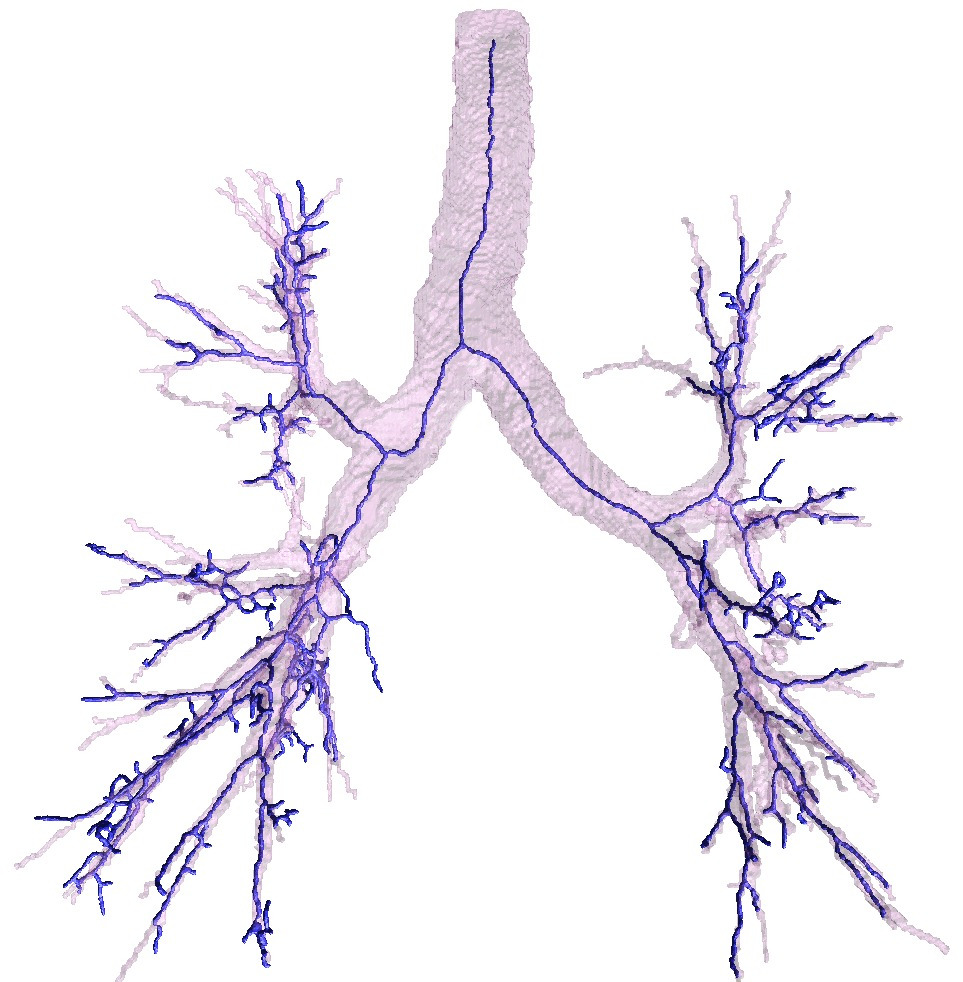}}\\
\vspace{0.3cm}
            (5) & (6) & (7) & (8) \\
      {\includegraphics[height=2.25cm]{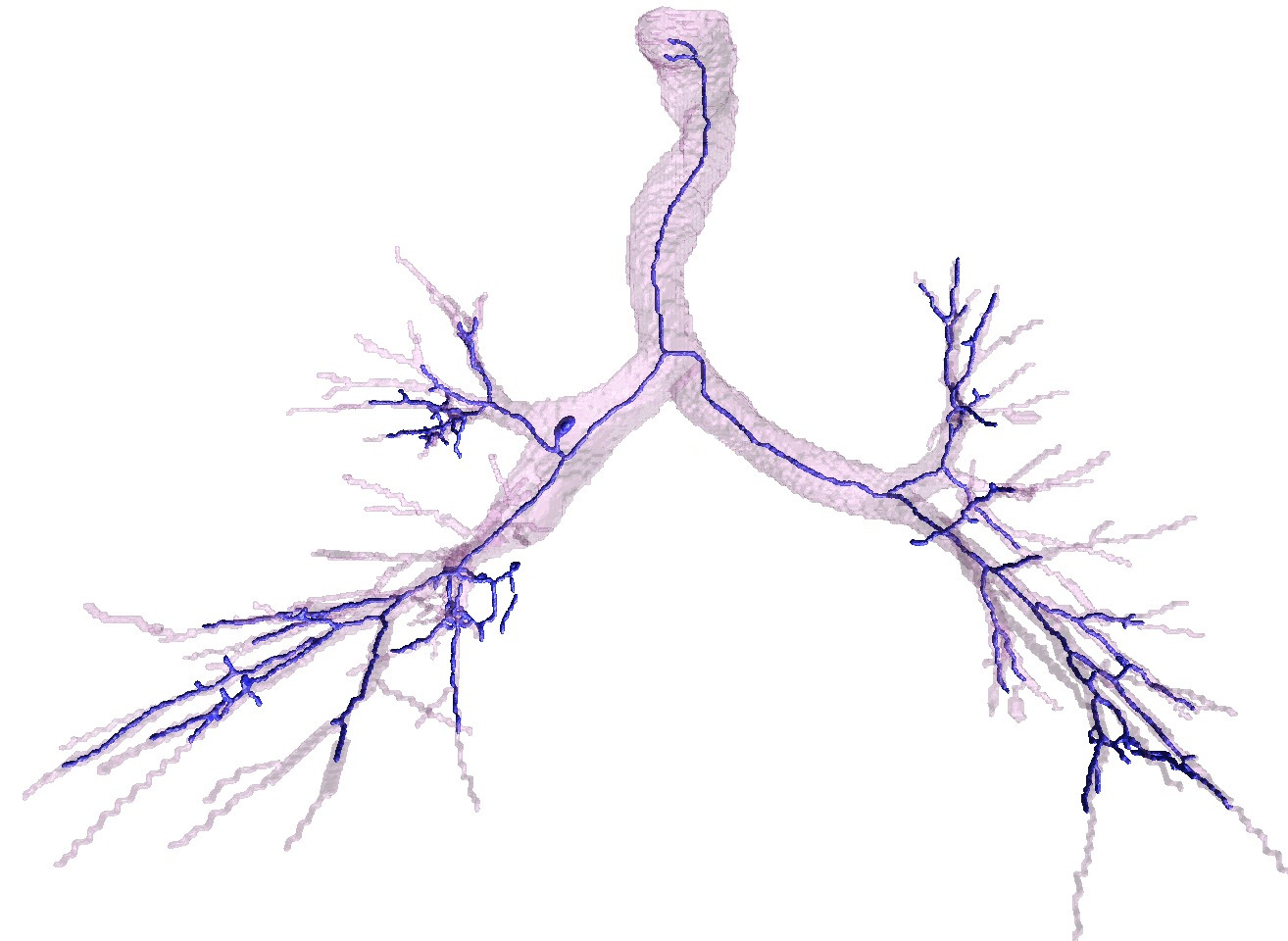}} &
      {\includegraphics[height=2.25cm]{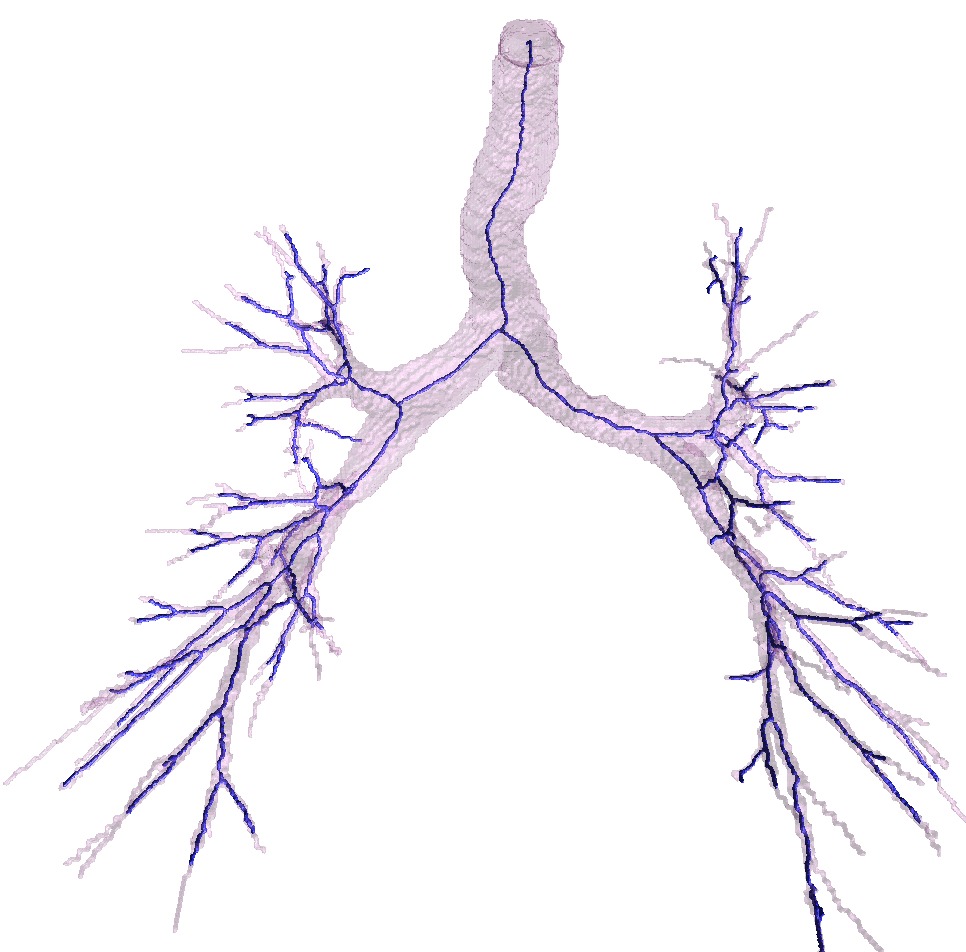}} & {\includegraphics[height=2.25cm]{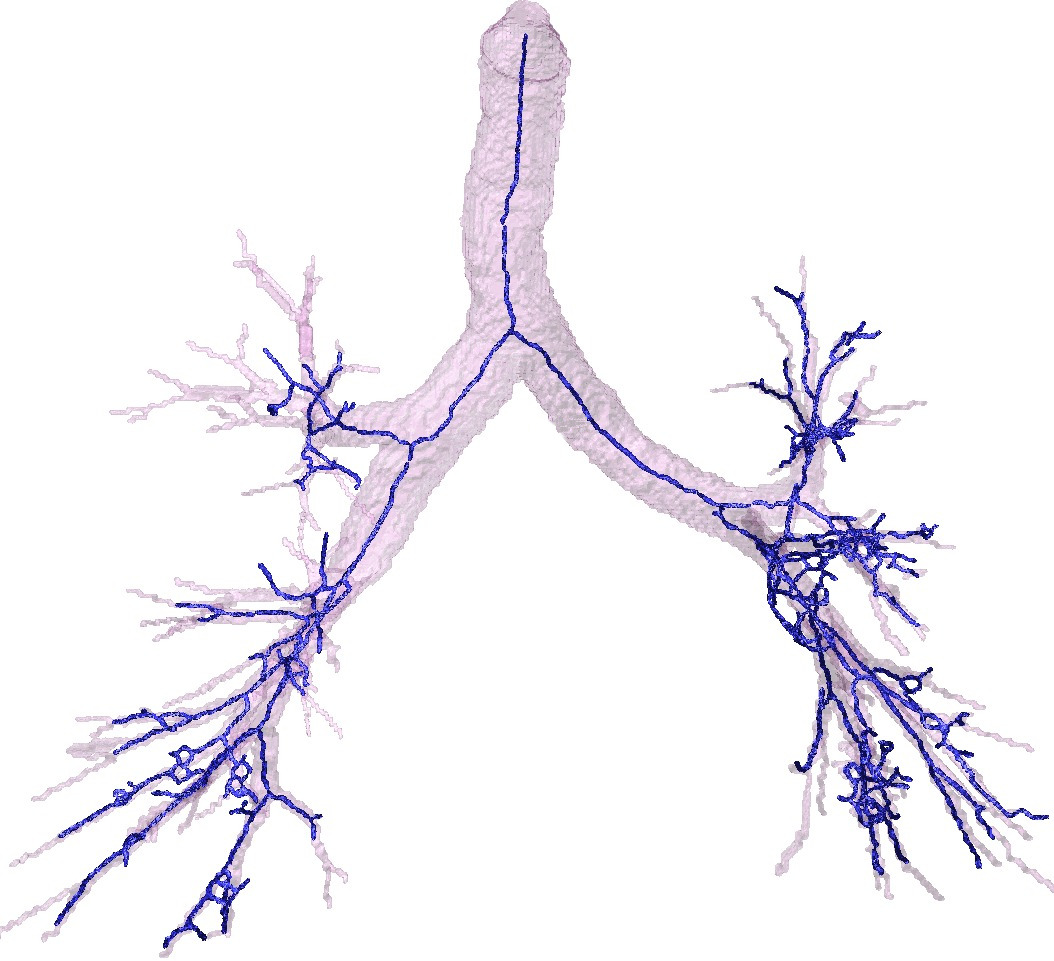}} & {\includegraphics[height=2.25cm]{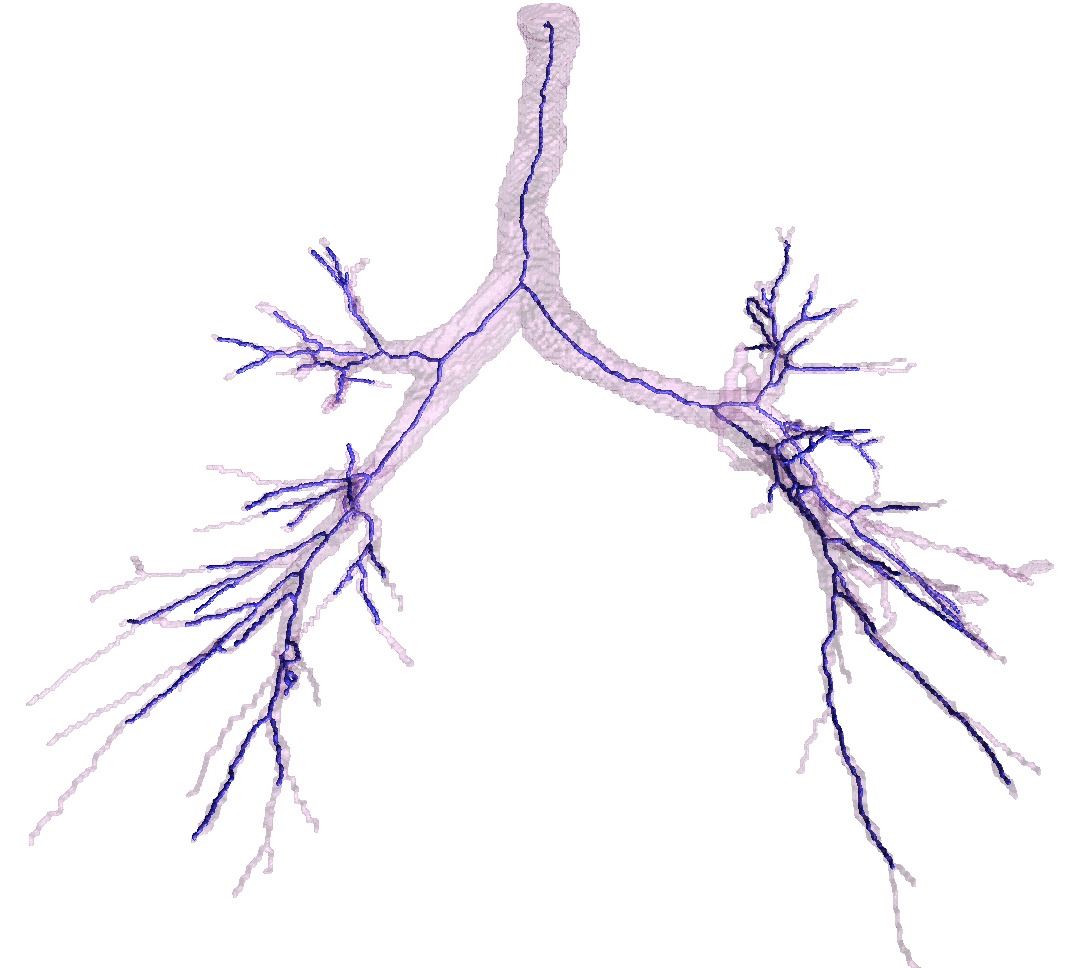}}\\
\vspace{0.3cm}
            (9) & (10) & (11) & (12) \\
      {\includegraphics[height=2.25cm]{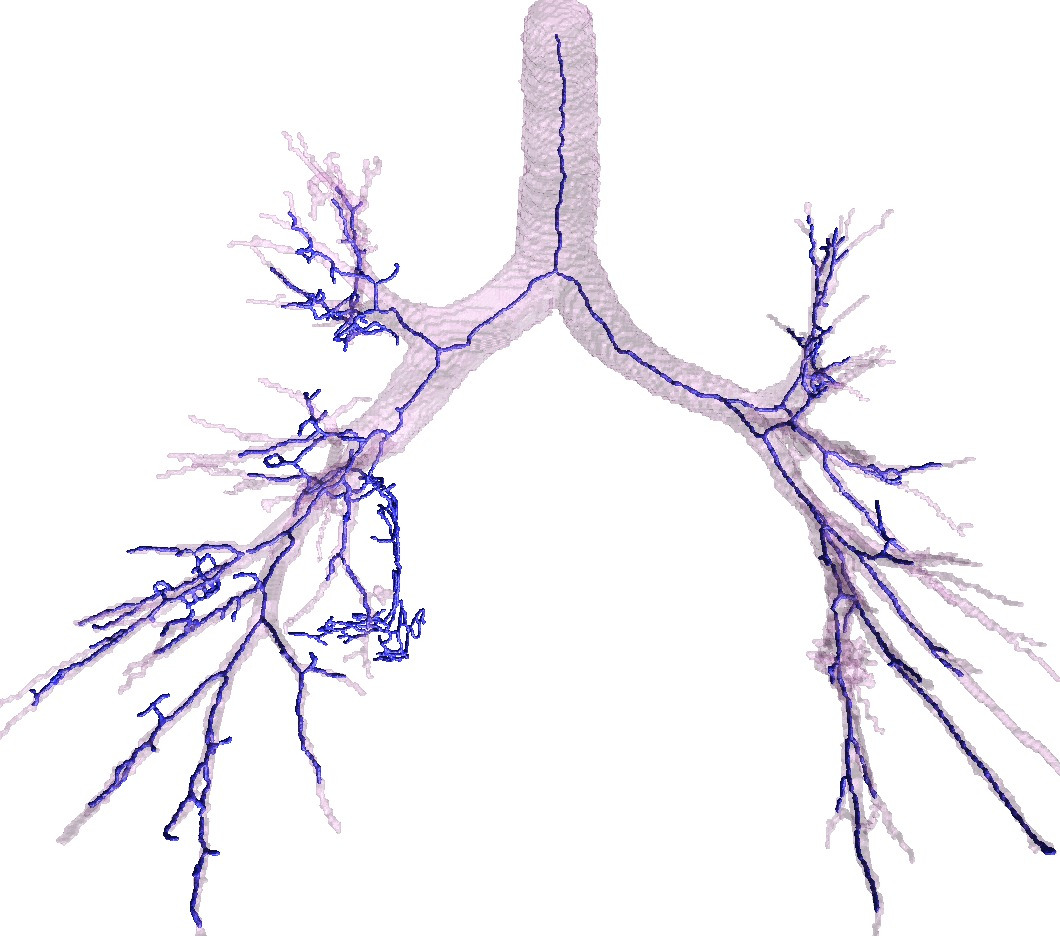}} &
      {\includegraphics[height=2.25cm]{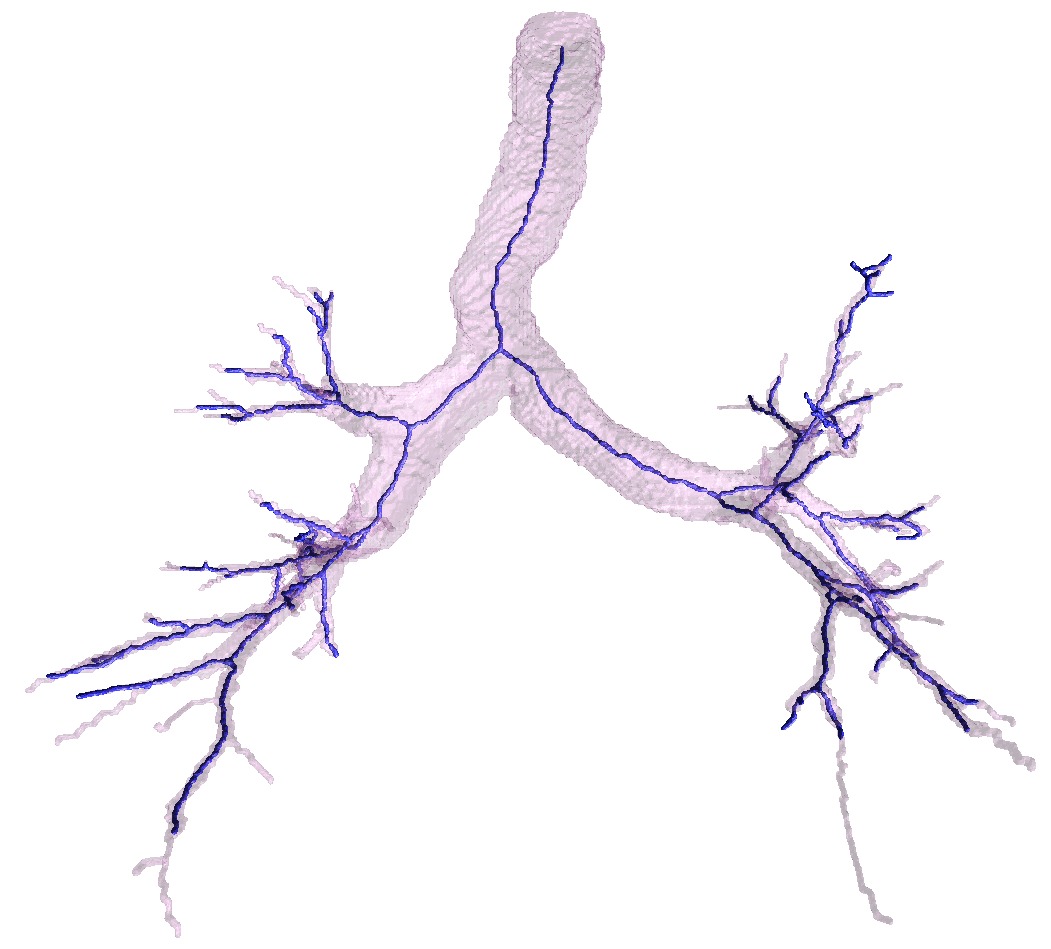}} & {\includegraphics[height=2.25cm]{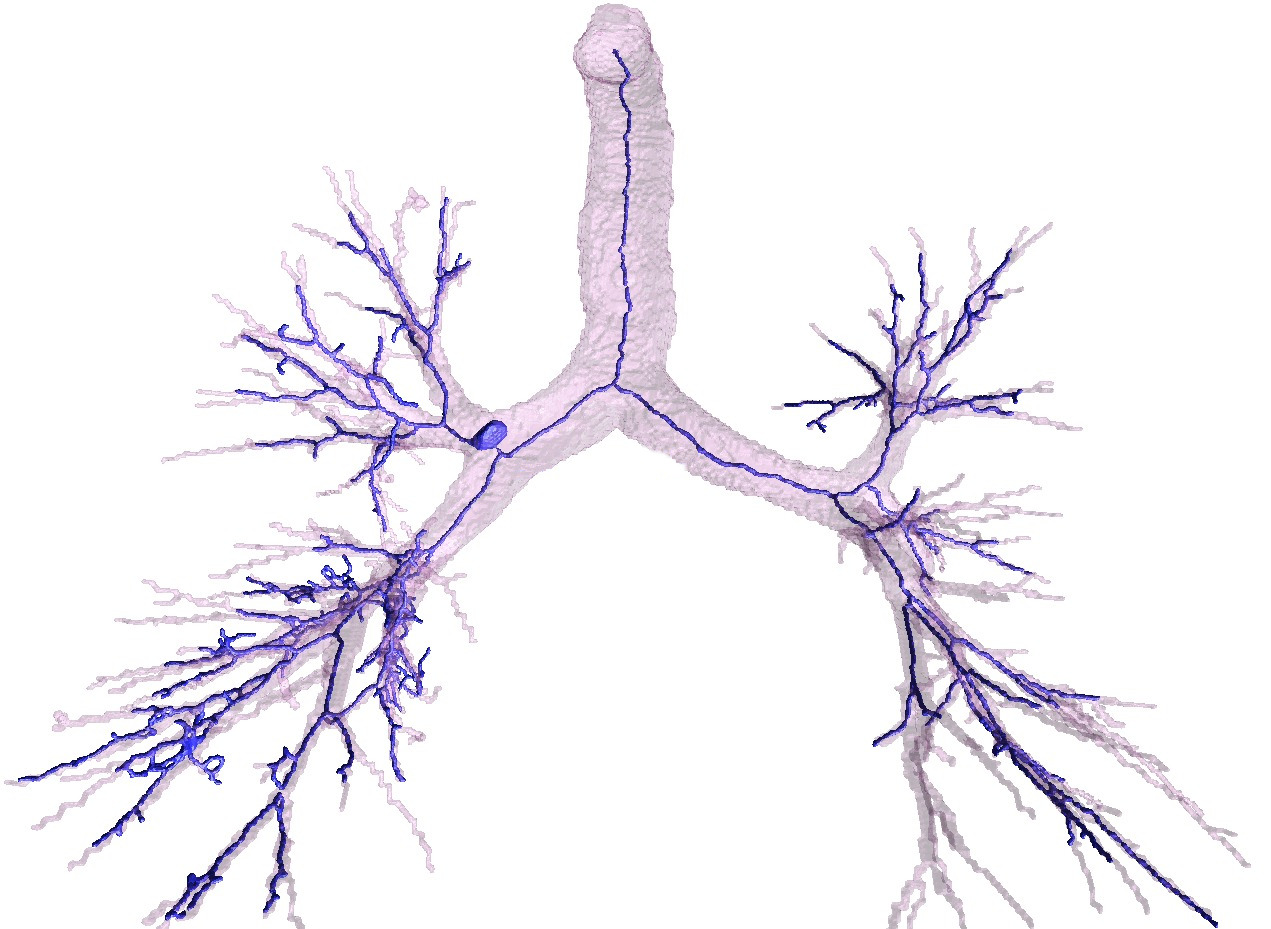}} & {\includegraphics[height=2.25cm]{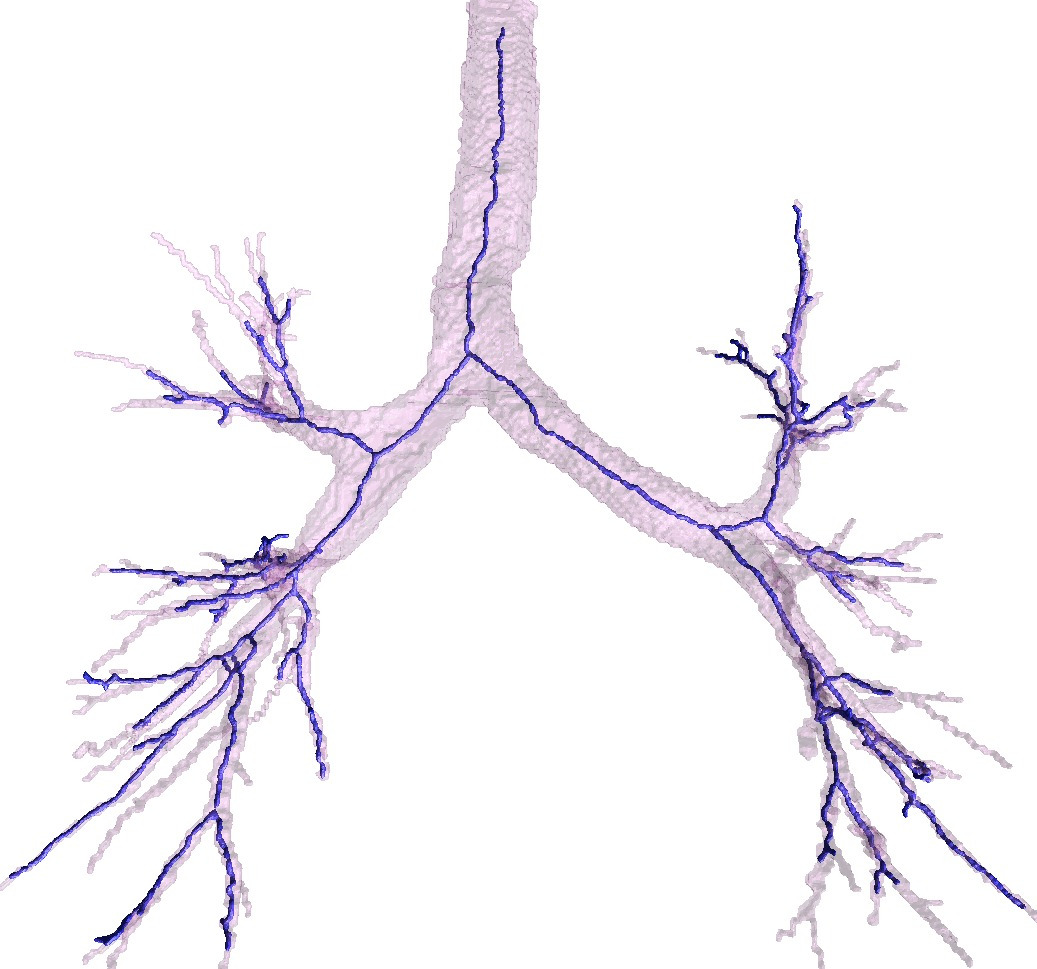}}\\
\vspace{0.3cm}
            (13) & (14) & (15) & (16) \\
\end{tabular}
\end{center}
\caption{Centerlines extracted from the segmentations obtained using the modified MHT method are shown embedded in the surface plot of the reference segmentations.} 
\label{fig:testPlots}
\end{figure}

The modified MHT method was evaluated on a test set, comprising of 16 images, and also compared with the original MHT, region growing on intensity and region growing on probability. The performance was measured by computing the error measure in~\eqref{eq:error}. 
The results obtained for the test set are presented in standard box-plots in Figure~\ref{fig:errPlot}. The modified MHT method clearly shows an improvement with respect to the original MHT method and region growing on intensity. It is, however, outperformed by region growing on probability images. 
The test set results for the modified MHT method are illustrated in Figure~\ref{fig:testPlots} along with the reference segmentations.

\section{Discussion and conclusion}

In this work, we looked at the possibility of using multiple hypothesis tracking for the purpose of automatic segmentation of airways from chest CT data. We discussed the limitations of the original MHT method in~\cite{friman}, which the presented work is based on and proposed modifications to it. We introduced statistical ranking of local hypotheses, which has significant impact on two important MHT parameters: local and global thresholds. Ranking altogether removes the need to threshold local hypotheses and makes the global threshold independent of the scale of airways. It also makes the global threshold less dependent on the particular fitness measure used in ranking the local hypotheses, as only the statistical significance of local hypotheses is used in making decisions. 
We also presented a different strategy to handle bifurcations which transfers the MHT tree state from the parent node to the new branch nodes. The effect of these modifications is that the modified MHT method can track airway branches of all sizes with the same set of parameters, and hence can track the entire airway tree starting from a single seed point, rendering the modified MHT method fully automatic. 

We evaluated the presented method and compared its performance with the original MHT method and region growing on intensity showing clear improvement in performance. We also compared it with region growing on probability, where the presented method does not outperform it. There are several aspects in which the presented method can be further improved starting with a strategy that adapts the number of local hypotheses dynamically depending on the surroundings instead of using a fixed number; this we expect will enable the method to detect more branches. Currently the evolution of the branches is not taken into account in any form. This can be altered by introducing a state space model that can capture the evolution of branches and predictions of local hypotheses can be based on it~\cite{statespace}. We only performed a coarse grid search of the tunable parameters; a finer and smarter search would most likely improve the results. The original MHT method in~\cite{friman} was shown to work well for vessel segmentation, albeit in a semi-automatic manner. With the modifications we have proposed, we expect the modified MHT method to be also applicable for automatic vessel segmentation and related applications. In conclusion, we do see further potential in using MHT in applications like segmentation which can benefit from tracking based on deferred decisions.

%
%

\end{document}